\documentclass{article}
\usepackage{ijcai19}

\usepackage{times}
\usepackage{soul}
\usepackage{url}
\usepackage[hidelinks]{hyperref}
\usepackage[utf8]{inputenc}
\usepackage[small]{caption}
\usepackage{graphicx}
\usepackage{amsmath}
\usepackage{booktabs}
\usepackage{algorithm}
\usepackage{algorithmic}
\usepackage{amssymb,amsfonts}
\usepackage{textcomp}
\usepackage{xcolor}
\usepackage{enumitem}
\usepackage{multicol}
\usepackage{multirow}
\usepackage[utf8]{inputenc}
\usepackage{hyperref}
\usepackage{subfigure}
\usepackage{comment}

\title{Accuracy Improvement of Neural Network Training using Particle Swarm Optimization and its Stability Analysis for Classification}

\author{
Arijit Nandi
\and
Nanda Dulal Jana
\affiliations
Department of Computer Science \& Engineering, National Institute of Technology Durgapur, Durgapur, India \\
\emails
an.17p10354@mtech.nitdgp.ac.in,
nandadulal@cse.nitdgp.ac.in
}

\begin{document}

\maketitle

\begin{abstract}
Supervised classification is the most active and emerging research trends in today's scenario. In this view, Artificial Neural Network (ANN) techniques have been widely employed and growing interest to the researchers day by day. ANN training aims to find the proper setting of parameters such as weights ($\textbf{W}$) and biases ($b$) to properly classify the given data samples. The training process is formulated in an error minimization problem which consists of many local optima in the search landscape. In this paper, an enhanced Particle Swarm Optimization is proposed to minimize the error function for classifying real-life data sets. A stability analysis is performed to establish the efficiency of the proposed method for improving classification accuracy. The performance measurement such as confusion matrix, $F$-measure and convergence graph indicates the significant improvement in the classification accuracy.

\end{abstract}

\section{Introduction}
Classification, is one of the most frequently encountered decision making tasks of human activity, problem occurs when an object needs to be assigned into a predefined group or class based on a number of observed attributes related to that object \cite{NNsurvey}. Example of classification includes medical diagnosis, quality control, speech recognition, classifying crops \cite{cropClassify}, pixel classification in satellite imagery \cite{AGRAWAL2015217}, forecasting oil demand in Iran \cite{oilIran}, classify multi spectral satellite image \cite{satImage2013}, hand written character recognition \cite{handWritten} etc. ANN is an important tool and widely used for classification. The advantage of ANN is, data driven self-adaptive because without any explicit functional or distributional details about the model they can adjust themselves accordingly. Second, ANN is $universal$ $approximators$ \cite{HORNIK1989} \cite{MALAKOOTI199827} where they can approximate functions which may be continuous and its counterpart discontinuous. Third, ANN is non linear in nature which makes them flexible to make complex real world relationships. Also they are comprehensive,tolerance to noisy data, parallelism and learning from example. Classification using ANN includes two parts- Learning(Training) of ANN and other is testing on unknown data samples for classification accuracy. The behaviour of neural network(NN) is affected by finding out the optimum NN architecture (i,e topology), transfer functions, initial weights and biases, training algorithm, epochs etc. But for achieving higher accuracy in prediction and classification NN should be trained properly. Fitness of the NN depends on initial weights and biases and the fitness value is minimum mean square error value given by NN while training. Number of $\textbf{W}$ and $b$ of the NN is decided by the NN topology. NN training is a costly and time consuming job. But why? Say, there are $n$ number of $\textbf{W}$ and $b$ required then there are $n!$ combinations of $\textbf{W}$, $b$ and $n!$ classification errors. Among those the minimum error will be considered and corresponding $\textbf{W}$ and $b$ combination will be considered for ANN testing. This is a computationally expensive problem. As the training process is formulated in an error minimization problem which consists of many local optima in the search landscape \cite{Malan2016}. Main objective is to find proper combination of $\textbf{W}$ and $b$ which will minimize the classification error. At a first go one can’t expect good result from the trained NN. The training process has to be done multiple times unless the expected output with minimum mean square error (MSE) is obtained. Or one can say that until the knowledge acquire is sufficient (i,e until the maximum iteration is reached or the a goal error value is achieved) the learning process should be repeated. Many deterministic (Back-Propagation \cite{Montana:1989}) and non-deterministic algorithms(Genetic Algorithm (GA)  \cite{Montana:1989}, Particle Swarm Optimization (PSO) \cite{nnPSO}, Artificial Bee Colony (ABC) Algorithm, Gravitational Search Algorithm(GSA) \cite{RASHEDI2011117}, hybrid PSO-GSA \cite{MIRJALILI201211125}, I-PSO-GSA \cite{HuCui}, Harmony search algorithm(HSA) \cite{HSAFFNN}) have been proposed and implemented to train NN. These algorithms have a tendency of less exploration (faster convergence and got stuck into local minima) and less exploitation ( more exploration i,e slow convergence) while training NN which affects the classification accuracy. This paper rectifies the problem by developing a new inertia weight strategy for PSO called PPSO which balances the exploration and exploitation properly while training ANN. By stability analysis it has been shown that the PPSO is stable. 
\par
The rest of the paper structure is as follows, Section 2 working principles of ANN, PSO. Section 3 discusses the proposed classification model. Section 4 is the training of ANN using PPSO. Experimental setup, result analysis and discussions are in section 6 and 7. Finally, Section 7 outlines the conclusions.
\section{Working Principles}
\subsection{Feed Forward Neural Network (FFNN)}
\begin{figure}[htbp]
\centering
\includegraphics[width=8cm, height=5cm]{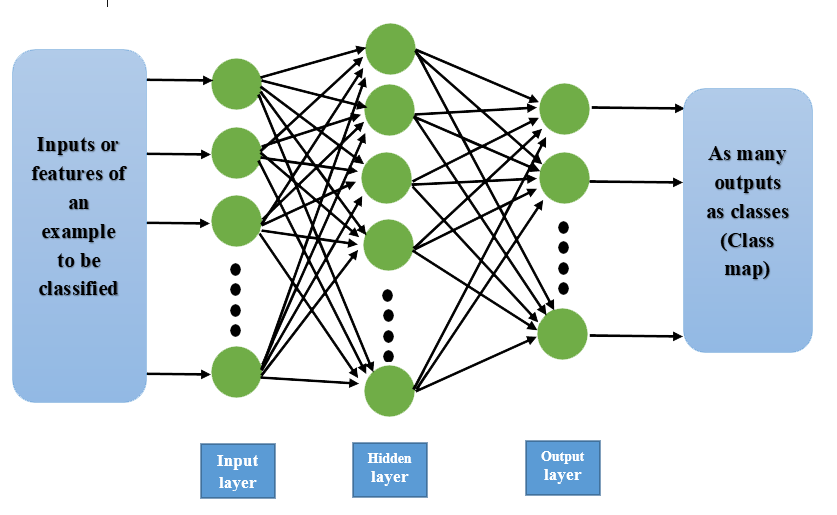}
\caption{Feed forward neural network}\label{fig:1}
\end{figure}

FFNN is, one of the kind of Multi-Layer Perceptron (MLP) ~\cite{MLP2007}, most popularly and most widely used model in many practical applications because of their high capability to forecasting and classification. MLP with three layers( First, input layer ($IL$); Second, hidden layer ($HL$); Last, output layer ($OL$)) is more popular \cite{Irie1988}. The inputs (n attributes) or features of an example to be classified are entered into the $IL$. The $OL$ typically consists of as many outputs as classes in classification problem. Information flows from $IL$ to $HL$ and then to $OL$ via a set of arcs \cite{Oza2012}. The basic computational element or building block of neural network is called neuron. It is to be noted that within the layer nodes are not directly connected. Figure. (\color{blue}\ref{fig:1}\color{black}) is an example. 
\par
The input to the hidden node j is obtained as
\begin{equation}
    {IH_j} = \sum\limits_{j = 1}^n {{W_{ij}}^{(1)}{O_i} + {b_j}}
    \label{equation:1}
\end{equation}
and output of hidden layer is
\begin{equation}
    {OH_j} = af(I{H_j})
    \label{equation:2}
\end{equation}
where $n$ is the number of neurons present in the $HL$. $ {W_{ij}}^{(k)} $  is the synaptic weight for the connection linking node $i$  in the ${K^{th} } $ layer of nodes to $j$. $af$  is the activation function (AF). Here the AF is considered as sigmoid function:
\begin{equation}
    af(I{H_j}) = \frac{1}{{1 + {e^{ - I{H_j}}}}}
    \label{equation:3}
\end{equation}
The input of an output node IO
\begin{equation}
    {IO_j} = \sum\limits_{i = 1}^z {{W_{ij}}^{(2)} * O{H_i}}
    \label{equation:4}
\end{equation}
and output of $OL$ is
\begin{equation}
    {OO_j} = af(I{O_j})
    \label{equation:5}
\end{equation}
where $z$ is number of hidden nodes. The outputs are non linearly related to inputs. 
The FFNN which is being considered for the problem of classification as pictured in Fig (\color{blue}\ref{fig:1}\color{black}). The range for output lie in [0, 1]. For each of the training examples in the training set($T_{set}$), its inputs are feeded to the FFNN and obtained (or predicted) outputs are calculated. The difference between each obtained (or predicted) and the corresponding targeted (or actual) output is calculated. The total error ($Err$) of the network is
\begin{equation}
    Err = \frac{1}{2}\sum\limits_{i = 1}^{|T_{set}|} {{{(T{O_i} - O{O_i})}^2}}
    \label{equation:6}
\end{equation}
where $TO_i$ and $OO_i$ are the targeted (or actual) and obtained (or predicted) outputs, respectively, for $i^{th}$ training example. Training of FFNN is nothing but minimizing the mean square error($MSE$). A separate error value (${Err}$) can be expressed as the parameters in the network as follows:
\begin{align}
    Err = \frac{1}{2}\sum\limits_{t = 1}^{|T_{set}|} {{{(T{O_i} - af(\sum\limits_{i = 1}^z {{W_{ij}}^{(2)}O{H_i}} ))}^2}}
    \nonumber \\
    = \frac{1}{2}\sum\limits_{t = 1}^{|T_{set}|} {{{(T{O_i} - af(\sum\limits_{i = 1}^z {{W_{ij}}^{(2)}af(\sum\limits_{j = 1}^n {{W_{ij}}^{(1)}{O_i} + {b_j}} )} ))}^2}} \nonumber
\end{align}

The learning algorithm typically repeated through $T_{set}$ many times- each repetition (or cycle) is called $epoch$ in the ANN literature.

\subsection{Particle Swarm Optimization (PSO)}
PSO is a stochastic population based meta-heuristic algorithm first introduced by Kennedy and Eberhart in 1995 ~\cite{Kennedy1995}. Suppose that the size of the swarm is $noP$ (population size) and the search space is $D$ - dimensional.
\par
The position of the $i^{th}$ particle is presented as $x_{id} = {\rm{ (}}x_{i1},x_{i2},.{\rm{ }}.{\rm{ }}.,x_{iD})$ where $x_{id} \in [l{b_d},u{b_d}] $, $d \in [1,D]$ and $lb_{d}$ and $ub_{d}$ are the lower and upper bounds of the $d^{th}$ dimension of the search space. The $i^{th}$ particle velocity is presented as $v_i = {\rm{ }}(v_{i1},v_{i2},.{\rm{ }}.{\rm{ }}.,v_{iD})$. At each time step, the position and velocity of the particles are updated according to the following equations:
  \begin{align}
  v_{ij}(t + 1) = \omega *v_{ij}(t) + {c_1}*{r_1}*\left( {{p^{lB}}_{ij}(t) - x_{ij}(t)} \right)
  \nonumber\\
  + {c_2}*{r_2}*\left( {{p^{gB}}_j(t) - x_{ij}(t)} \right)
     \label{equation:7}
\end{align}

\begin{equation}
    x_{ij}(t + 1) = x_{ij}(t) + v_{ij}(t + 1)  
    \label{equation:8}
\end{equation}
where ${r_1},{r_2}$ are two distinct random numbers, generated uniformly from the range [0, 1], ${c_1},{c_2}$ are acceleration coefficients and $t$ is the current iteration. The best previous position found so far by the particle is denoted as ${p^{lB}_i}{\rm{ }}$, and the best previous position discovered by the whole swarm is denoted as ${p^{gB}}{\rm{ }}$. The velocity of particle should be under the constrained conditions ${\left[ {{v_{min}},{\rm{ }}{v_{max}}} \right]^D}$. In Eq. (9), ${\omega} *v_{ij}(t)$ , provides exploration ability for PSO. The second part ${c_1}*{r_1}*\left( {{p^{lB}}_{ij}(t) - x_{ij}(t)} \right)$ and third part ${c_2}*{r_2}*\left( {{p^{gB}}_j(t) - x_{ij}(t)} \right)$ represent private thinking and collaboration of particles respectively.
\par
It has been proved by experiments that truly random initialization of particles position in search space can enhance the exploration of various regions in search space and can also enhance the performance of the PSO. Hence classification performance of FFNN will be increased. However computers can’t provide truly random numbers via pseudo random generators, which means optimal discrepancy is not achievable by them.
\par
\begin{figure}[htbp]
\includegraphics[width=8cm, height=3cm]{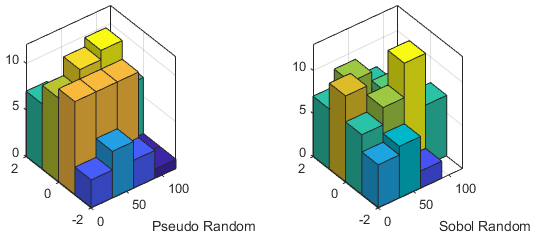}
\caption{Pseudo random (left) and Sobol random (right) initialization in PSO position}\label{fig:2}
\end{figure}

Figure. (\color{blue}\ref{fig:2}\color{black}) shows the difference between pseudo random and Sobol random initialization.
\par
\par
So, many researchers have studied alternative ways to generate low-discrepancy sequences. \cite{Uy2007} have investigated some of well-known randomized low-discrepancy sequences (Sobol, Halton and Faure) for the PSO position initialization. So, according to the research, PSO initialization with Sobol sequences is recommended \cite{Jordehi2013}.
\par
\cite{shi2001} claimed that a large $IW$ facilitates a global search while a small $IW$ facilitates a local search.

\section{Proposed  Classification Model}
\par
FFNN with three layers is used in this experiment to perform the classification problems. For higher classification accuracy, efficiency the weights and biases of the FFNN should be initialized efficiently before performing the classification task and prediction. For optimizing the FFNN's weights and biases the PPSO is used. Fitness function of PPSO is MSE as follows:
\begin{equation}
    MSE = \frac{1}{{{T_{set}}}}\sum\limits_{t = 1}^{|{T_{^{set}}}|} {\sum\limits_{i = 1}^n {{{(T{O_i}^t - O{O_i}^t)}^2}} }
    \label{equation:10}
\end{equation}
where, $t$ is number of training sample , ${OO}_i^t$ is desired output and ${TO}_i^t$ is actual output of the $i^{th}$ input in the $t^{th}$ training sample respectively.
\par
In PPSO particle initial positions have been initialized by Sobol random initializer instead of pseudo random initializer.We have proposed a new hyperbolic tangent increasing inertia weight strategy. The formula can be defined as follows:
\begin{equation}
    {\omega} (t) = {\omega _{\min }} + \tanh ({I_{current}} \times {\rm{((}}{\omega _{\max }}{\rm{ - }}{\omega _{\min }}{\rm{)/}}{{\rm{I}}_{\max }}{\rm{)}})
    \label{equation:11}
\end{equation}
where ${\omega}_{min}$ is initial $IW$ and ${\omega}_{max}$ is final $IW$ . $I_{max}$ is maximum iteration number. $I_{current}$ is the current iteration number. ${\omega}{(t)}$ is $IW$ for PPSO  which is	time varying.
\par
Suppose that the structure of the FFNN is a ${p-q-r}$ structure, where $p$ is number of nodes present in $IL$, $q$ is number of nodes present in $HL$ and $r$ is number of nodes present in $OL$. And there are ${noP}$ number of particles in the population, where every particle $x_i\left( {i = 1,{\rm{ }}2,{\rm{ }}3,{\rm{ }}4,{\rm{ }} \ldots  \ldots .,noP} \right)$ is a $D$-dimensional vector ${\left( {x_{i1},{\rm{ }}x_{i2},{\rm{ }}x_{i3},{\rm{ }} \ldots ..,{\rm{ }}x_{iD}} \right)}$ where ${D=nw+nb}$ where $nw$ is total number of weights for the links, $nb$ is total number of biases present in $HL$ and $OL$. Now $nw=pq+qr$ and $nb=q+r$. Therefore $D=pq+qr+q+r$. We have considered $p_i$ into the weights and biases of FFNN, where the components $x_{i,1},{\rm{ }}x_{i,2},{\rm{ }} \ldots  \ldots .,{\rm{ }}x_{i,pq}$ of  are weights between $IL$ and $HL$, the components $x_{i,pq + 1},{\rm{ }}x_{i,2},{\rm{ }} \ldots  \ldots .,{\rm{ }}x_{i,pq + q}$  of  are the biases of $HL$, the components $x_{i,pq + q + 1}, \ldots  \ldots .,{\rm{ }}x_{i,pq + q + qr}$ of $x_i$ are the weights between $HL$ and $OL$, and the components $x_{i,pq + q + qr + 1}, \ldots  \ldots .,{\rm{ }}x_{iD}$ of $x_i$ are the biases of $OL$.
\section{Training of FFNN using PPSO}
\par
The following steps are for the proposed algorithm:
\begin{itemize}
    \item { \textbf{1:} Set control parameters of the proposed PPSO algorithm.}\
    
    \item {\textbf{2:} Initialize position $x_i\left( {i = 1,{\rm{ }}2,{\rm{ }}3,{\rm{ }}4,{\rm{ }} \ldots  \ldots .,noP} \right)$ of particle using Sobol random initializer.}\
    
    \item {\textbf{3:} Map $x_i$  into weights and biases of FFNN and take each tuple from the dataset and train the FFNN using Equation. (\color{blue}\ref{equation:1}\color{black}), eq. (\color{blue}\ref{equation:2}\color{black}), eq. (\color{blue}\ref{equation:4}\color{black}), and eq. (\color{blue}\ref{equation:5}\color{black}).This phase is called training of FFNN.}\
   
    \item {\textbf{4:} Fitness (i,e FFNN error or MSE) can be obtained using eq. (\color{blue}\ref{equation:10}\color{black}). }\
    
    \item {\textbf{5:} Now calculate IW by using eq. (\color{blue}\ref{equation:11}\color{black}).}\
    
    \item {\textbf{6:} Update velocity $v_{ij}{(t + 1)}$ and positions $x_{ij}(t + 1)$ of particles according to eq. (\color{blue}\ref{equation:7}\color{black}) and eq. (\color{blue}\ref{equation:8}\color{black}) respectively. }\
    
    \item {\textbf{7:} Go to \textbf{8} if stopping criteria meets. Else, \textbf{3}. }\
    
    \item {\textbf{8:} Output the best particle that is mapped into weights and biases of FFNN. The initial parameters (weight and bias) are obtained using PPSO. Then test the trained FFNN. }\
\end{itemize}
\section{Experimental Setup }
\subsection{Dataset Description}
In this experiment, 8 different datasets are used. All are collected from UCI repository ~\cite{Dua:2017}. Data normalization is done in the range between [-1, 1] and there is no missing value present in dataset. The details of data sets are mentioned TABLE (\color{blue}\ref{table:I}\color{black}):
\begin{table}[htbp]
\centering
\resizebox{0.4\textwidth}{!}{\begin{tabular}{llll} 
\hline
Dataset Name & Instances & Attributes & Classes \\ 
\hline
Wine & 178 & 13 & 3 \\
Iris & 150 & 4 & 3 \\
Breast Cancer & 569 & 31 & 2 \\
Banknote & 1372 & 5 & 2 \\
Balance Scale & 625 & 4 & 3 \\
Appendicitis & 106 & 7 & 2 \\
Thyroid gland & 215 & 5 & 3 \\
Ionosphere & 351 & 34 & 2 \\
\hline
\end{tabular}}
\caption{DETAILS OF 8 DATASETS USED IN THIS PAPER}
\label{table:I}
\end{table}
\par
 “Hold Out” cross validation method has been used for the data division where 80$\%$ instances of data are used for training the FFNN and rest 20$\%$ instances of data are used to test the trained FFNN.
\subsection{Parameter setup for algorithms }
In this experiment, the population size (${noP}$) is 50, $I_{max}$ is 500, and initialization of $p_i$ and $v_i$ are initialized by Sobol random initializer. In PPSO $c_1$ and $c_2$ are 1.6 and 1.7 respectively and $IW$ increases linearly from 0.4 to  0.9. In BPSO $c_1$and $c_2$ both are set to 1.5 and $IW$ is varying inertia with value is decreasing from ${\omega}_{max}$ ${=0.9}$ to ${\omega}_{max}$ ${=0.3}$. In SGPSO, $c_1$,$c_2$ and $IW$ all are same as BPSO and $c_3$ is 0.5. For SGPSO geometric center is 100 \cite{Beatriz2015}. In PSOGSA $c_1$ and $c_2$ both are set to 1 and $IW$ linearly decreases from ${\omega}_{max}$ ${=0.9}$ to ${\omega}_{max}$  ${=0.5}$. In GSA, gravitational constant value ( $G_0$) is set to 1 and initial values for $mass$ and $acceleration$ are set to 0. 
\subsection{Performance metric}
For the classification accuracy (testing accuracy) measurement Confusion Matrix is used. Confusion Matrix is very useful tool in order to analyze the accuracy \cite{Han2011Data}, that how efficiently the classifier can able to recognize different classes. The classifier is also evaluated using standard metrics: Precision (Pre), Recall (Rec), and F-Measure (FM) [\ref{equation:12}] \cite{Aggarwal} :
\begin{equation}
    FM = 2*(\frac{{Pre*Rec}}{{Pre + Rec}})
    \label{equation:12}
\end{equation}
where $Pre = \frac{{Tp}}{{Tp + Fp}}$ and $Rec = \frac{{Tp}}{{Tp + Fn}}$.
True positives ($Tp$): number of incorrect samples correctly identified as incorrect; true negatives ($Tn$): number of normal samples correctly identified as normal; false positives ($Fp$): number of normal samples misclassified as incorrect; false negatives ($Fn$): number of incorrect samples misclassified as normal. Here FM is the weighted average of Precision and Recall. Therefore, this score takes both false positives and false negatives into account.
\section{Analysis and Discussion}

\subsection{Stability Analysis of PPSO}
From eq.\ref{equation:7} and \ref{equation:8} each dimension is updated independently from other dimensions. The relation between the dimensions and problem space is the global best position ${p^{gB}}$   found so far. So, without loss of generality, the algorithm can be reduced to one dimensional case for analysing purpose: 
\begin{equation}
v(t + 1) = \omega *v(t) + \psi_1*\left( {{p^{lB}} - x(t)} \right)
  + \psi_2*\left( {{p^{gB}} - x(t)} \right) 
  \label{equation:13}
\end{equation}
  
\begin{equation}
    x(t + 1) = x(t) + v(t + 1)  
    \label{equation:14}
\end{equation}
where $\psi_1$=$c_1*r_1$ and $\psi_2$=$c_2*r_2$. \\
Let 
\begin{align}
    \psi =\psi_1+\psi_2 \nonumber \\
    p=\frac{\psi_1p^{lB}+\psi_2p^{gB}}{\psi} \nonumber
\end{align}
Then Eq.\ref{equation:13} and \ref{equation:14} can be simplified as 
\begin{equation}
  v(t + 1) = \omega v(t) - \psi x(t)+ \psi p
  \label{equation:15}
\end{equation}
\begin{equation}
    x(t + 1) = \omega v(t) +(1-\psi) x(t)+\psi p 
    \label{equation:16}
\end{equation}
where $\psi$ is the new attraction coefficient, which is the combination of cognitive behaviour (local search) and  social behaviour (global attraction) $\psi1$ and $\psi2$. The attraction point($p$) is weighted average of global best($p^{gB}$) and local best ($p^{lB}$) in the swarm. \\
\textbf{Dynamic Analysis:}
For dynamic analysis ~\cite{MClerc2002}, let 
\begin{align}
    y_t=x_t-p
\end{align}
Now Eq. \ref{equation:15} and \ref{equation:16} can be simplified as:
\begin{equation}
  v(t + 1) = \omega v(t) - \psi y(t)
  \label{equation:17}
\end{equation}
\begin{equation}
    y(t + 1) = \omega v(t) +(1-\psi) y(t) 
    \label{equation:18}
\end{equation} \\
Let 
\begin{equation}
    {X_t} = \left( \begin{array}{l}
{v_t}\\
{x_t}
\end{array} \right)
\label{equation:19}
\end{equation}
then Eq.(\ref{equation:17}) and \ref{equation:18} can be written in matrix-vector form:
\begin{equation}
    X_{t+1}=GX_{t}
\end{equation}
so, 
\begin{equation}
    \left( \begin{array}{l}
{v_{t + 1}}\\
{x_{t + 1}}
\end{array} \right) = \left( {\begin{array}{*{20}{c}}
\omega &{ - \psi }\\
\omega &{1 - \psi }
\end{array}} \right)\left( \begin{array}{l}
{v_t}\\
{x_t}
\end{array} \right)
\label{equation:21}
\end{equation}
where the coefficient matrix is $G$
\begin{align}
G=\left( {\begin{array}{*{20}{c}}
\omega &{ - \psi }\\
\omega &{1 - \psi }
\end{array}} \right)
\end{align}
\par 
The eigenvalues of the $G$ can be obtained by solving the following equation for  $\lambda$ :
\begin{equation}
\det\begin{pmatrix} \omega -\lambda & - \psi \\\omega & (1- \psi)-\lambda\end{pmatrix} =0
\end{equation}

\[ \Rightarrow (\omega  - \lambda )(1 - \psi  - \lambda ) + \omega \psi  = 0 \]
\[ \Rightarrow {\lambda ^2} - \lambda (1 - \psi ) - \omega \lambda  + (\omega (1 - \psi ) + \omega \psi ) = 0 \]
\[ \Rightarrow {\lambda ^2} - \lambda ((1 - \psi ) + \omega ) + \det (G) = 0\]
\begin{align}
 \Rightarrow {\lambda ^2} - Tr(G) \lambda + \det (G) = 0
 \label{equation:24}
\end{align}

where  $\det (G)$ is the determinant of coefficient matrix $G$. 
\[ \det (G)= det \left( {\begin{array}{*{20}{c}}
\omega &{ - \psi }\\
\omega &{1 - \psi }
\end{array}} \right)= \omega\]
$Tr(G)$ is trace of matrix $G$ which is equal to the summation of principal diagonal elements. Here \[ Tr(G) = \omega +1-\psi \]

Eq. \ref{equation:24} is the characteristic equation. Where $\lambda$ is the eigen value which is a solution to the eq \ref{equation:24}. 
Now 
\[\lambda  = \frac{{Tr(G) \pm \sqrt {Tr{{(G)}^2} - 4del(G)} }}{2}\]
Now for checking the condition in which the system is stabile. The real part of eigen value ( $Re(\lambda)$ ) can be given as follows:

\[{\mathop{\rm Re}\nolimits} (\lambda ) = \left\{ \begin{array}{l}
\frac{{Tr(G)}}{2} \hspace{0.5cm} if \hspace{0.5cm} Tr{(G)^2} < 4\det (G)\\
\frac{{Tr(G) + \sqrt {Tr{{(G)}^2} - 4\det (G)} }}{2} \hspace{0.2cm} if \hspace{0.2cm} Tr{(G)^2} \ge 4\det (G)
\end{array} \right.\]

If $Tr{(G)^2} < 4\det (G)$, the stability condition is as follows:
\begin{align}
Tr(G)< 0
\label{equation:25}
\end{align}
If $Tr{(G)^2} \ge 4\det (G)$, the stability condition can be derived as follows:
\[{Tr(G) + \sqrt {Tr{{(G)}^2} - 4\det (G)} } <0\] 

\[\Rightarrow {\sqrt {Tr{{(G)}^2} - 4\det (G)} } < -Tr(G)\]

squaring bothsides the following can be achieved
\[ \Rightarrow {\sqrt {Tr{{(G)}^2} - 4\det (G)} } < Tr(G)^2 \]
so, 
\begin{align}
det(G) >0
\label{equation:26}
\end{align}
 
 By substituting  $Tr(G) = \omega +1-\psi $ to eq.(\ref{equation:25}) the follwing can be concluded 
 \[ \omega +1-\psi < 0\]
 \begin{align}
    \Rightarrow \omega < \psi-1 
 \end{align}
 By substituting  $det(G) = \omega $ to eq.(\ref{equation:26}) the follwing can be concluded that 
 \begin{align}
    \omega >0
 \end{align}
 By combining all the results above, the conclusion can be drawn that the proposed PPSO is stable and its stability depends on $\omega$. The range of $\omega$ for which the PPSO is stable is $0< \omega <(\psi-1)$. Hence the proposed PPSO is stable.

\subsection{Experimental Analysis}
The $F$-measure and accuracy from best run are reported in TABLE (\color{blue} \ref{table:III} \color{black}) and TABLE (\color{blue}\ref{table:II}\color{black}) respectively for BPSO, SGPSO, GSA, PSOGSA and PPSO classifiers. The results in boldface in tables are showing better in comparative analysis of four above mentioned classifiers.
\begin{table}[htbp]
\centering
\resizebox{0.45\textwidth}{!}{\begin{tabular}{llllll} 
\hline
Dataset & BPSO & SGPSO & GSA & PSOGSA & PPSO \\ 
\hline
Wine & 88.89 & 96.29 & 96.3 & 96.2 & \textbf{100}  \\
Iris & 96.96 & 93.93 & 96.9 & 100 & 100 \\
Breast Cancer & 89.5 & 87.71 & 92.1 & 93.9 & \textbf{95.6}  \\
Banknote & 98.5 & 98.5 & 98.5 & 98.9 & \textbf{99.63}  \\
Balance Scale & 88.70 & 88.7 & 87 & 90.3 & \textbf{93.5}  \\
Appendicitis & 90.9 & 90.9 & 90.9 & 90.9 & \textbf{100}  \\
Thyroid gland & 93.8 & 84.37 & 84.4 & 90.6 & \textbf{96.9}  \\
Ionosphere & 92.45 & 90.56 & 84.9 & 90.6 & \textbf{96.2}  \\
\hline
\end{tabular}}
\caption{Accuracy comparison for the PPSO classifier with other classifiers (BPSO, SGPSO, GSA, PSOGSA)}
\label{table:II}
\end{table}
\par
Table (\color{blue}\ref{table:II}\color{black}), reports the accuracy comparison of the proposed PPSO classifier with 4 other classifiers (i,e BPSO, SGPSO, GSA, PSOGSA). It can be reported that on all datasets PPSO has outperformed BPSO, SGPSO, GSA.And PPSO outperformed PSOGSA in 7 datasets out of 8 datasets and  achieved better classification accuracy than PSOGSA.
But there is a tie between PSOGSA and PPSO, both have achieved 100\% classification accuracy on Iris dataset.
\par
From TABLE (\color{blue}\ref{table:III}\color{black}), the PPSO classifier showed better performance in 7 out of 8 datasets. But PSOGSA classifier, among 8 datasets 1 dataset (i,e Balance Scale) has shown better FM value than PPSO.

\begin{table}[htbp]
\centering
\resizebox{0.45\textwidth}{!}{\begin{tabular}{llllll} 
\hline
Dataset & BPSO & SGPSO & GSA & PSOGSA & PPSO \\ 
\hline
Wine & 86.9 & 95.23 & 86.9 & 95.2 & \textbf{100}  \\
Iris & 94.7 & 90 & 94.7 & 100 & 100 \\
Breast Cancer & 86 & 84.7 & 88.17 & 93.4 & \textbf{95.15}  \\
Banknote & 98.7 & 98.72 & 98.7 & 99.04 & \textbf{99.67}  \\
Balance Scale & 87.27 & 87.27 & 85.7 & 89.6 & 88.9 \\
Appendicitis & 66.66 & 80 & 66.66 & 80 & \textbf{100}  \\
Thyroid gland & 94.7 & 88.9 & 88.8 & 93.2 & \textbf{97.43}  \\
Ionosphere & 94.7 & 93.33 & 88.8 & 92.3 & \textbf{97.29}  \\
\hline
\end{tabular}}
\caption{F measure comparison for the PPSO classifier with other classifiers (BPSO, SGPSO, GSA, PSOGSA)}
\label{table:III}
\end{table}
\par
\begin{figure}[htbp]
\centering
\includegraphics[width=6cm, height=4cm]{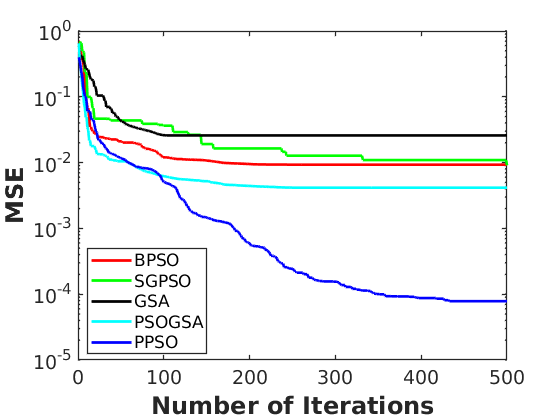}
\caption{Convergence curves of BPSO, SGPSO, GSA, PSOGSA and PPSO for banknote dataset classification}\label{fig:3}
\end{figure}

Figure (\color{blue}\ref{fig:3}\color{black}) reports the convergence curves while training the FFNN using BPSO, SGPSO, GSA, PSOGSA and PPSO. One dataset ( i,e banknote dataset) is reported for the convergence curve plot. It can be seen that the convergence of PPSO is better than other classifiers.
\par
\par
\begin{figure}[htbp]
\centering
    \subfigure[]{\label{fig:a}\includegraphics[width=0.32\linewidth]{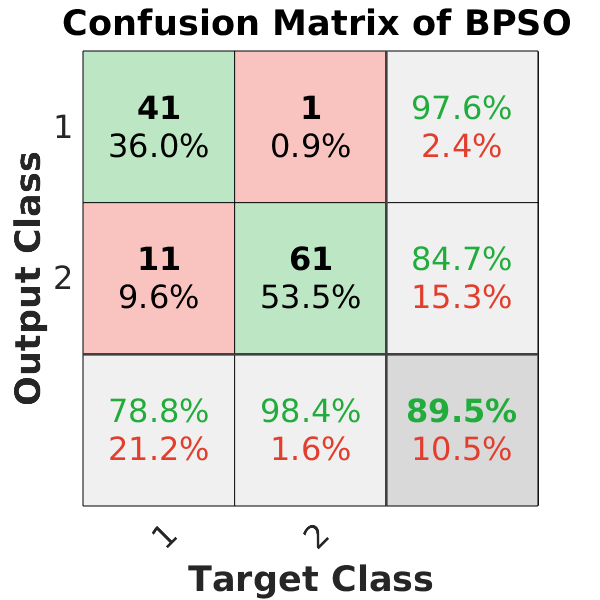}}
    \subfigure[]{\label{fig:b}\includegraphics[width=0.32\linewidth]{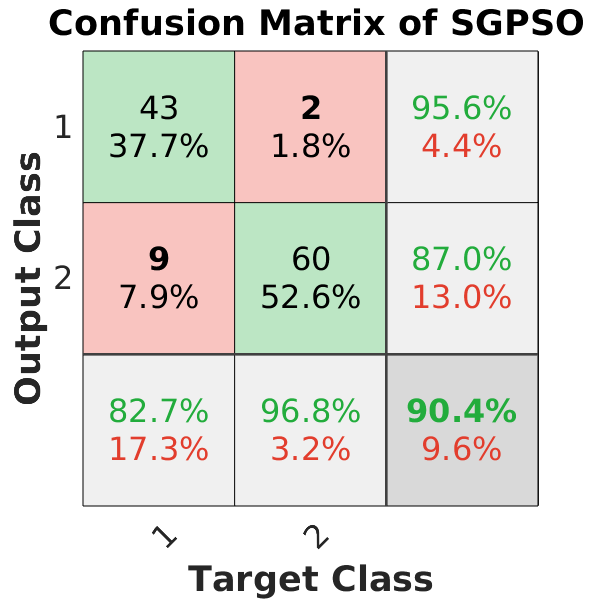}} 
    \subfigure[]{\label{fig:c}\includegraphics[width=0.32\linewidth]{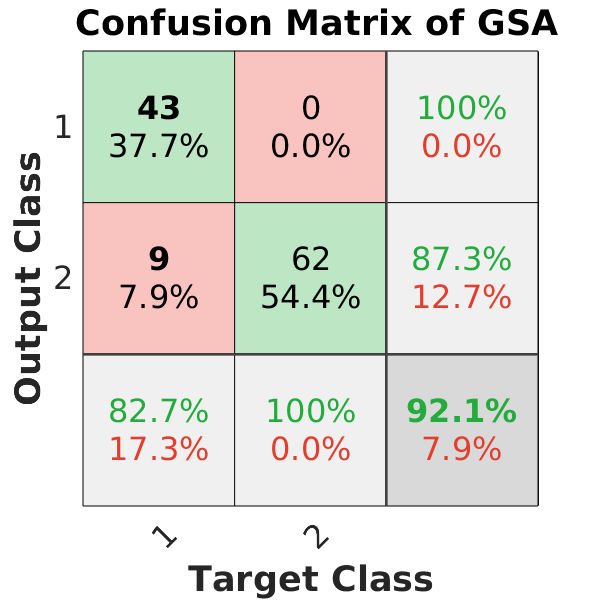}} \\
    \subfigure[]{\label{fig:d}\includegraphics[width=0.32\linewidth]{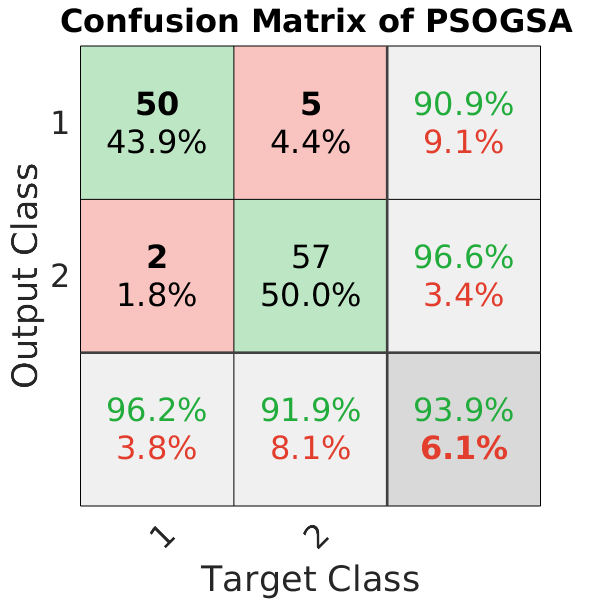}} 
    \subfigure[]{\label{fig:e}\includegraphics[width=0.32\linewidth]{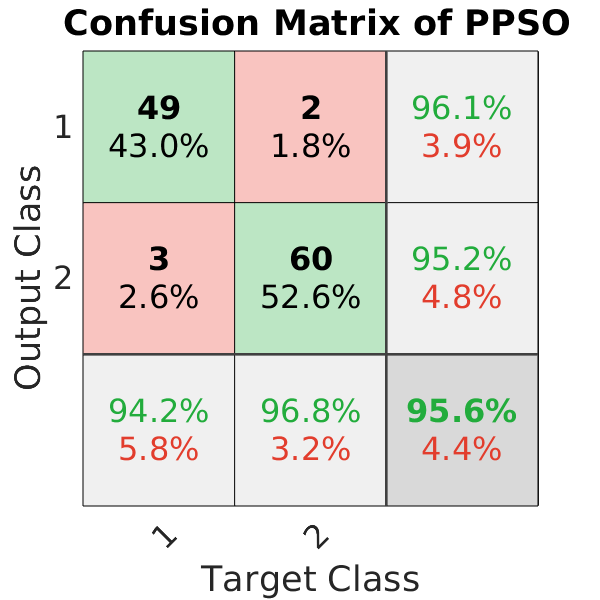}}
  \caption{Confusion matrix comparison among 5 classifiers \ref{fig:a} BPSO, \ref{fig:b} SGPSO, \ref{fig:c} GSA, \ref{fig:d} PSOGSA, \ref{fig:e} PPSO}
  \label{fig:4}
\end{figure}

Figure (\color{blue}\ref{fig:4}\color{black}) reports the confusion matrix (a measurement between target class(actual class) vs output class (FFNN produced class)) comparison among 5 classifiers (BPSO, SGPSO, GSA, PSOGSA and PPSO) for classification of  cancer dataset. In confusion matrix plot green colored cells are the classes which are correctly classified and red colored cells are the classes  which are misclassified by the classifier for the cancer test dataset. It also shows the accuracy of all the classifiers. Subfigure \color{blue}\ref{fig:a}\color{black}, \color{blue}\ref{fig:b}\color{black}, \color{blue}\ref{fig:c}\color{black}, \color{blue}\ref{fig:d}\color{black}, \color{blue}\ref{fig:e}\color{black} \quad are the confusion matrix plot for BPSO, SGPSO, GSA, PSOGSA, PPSO classifier's respectively. And the difference among classifiers are clearly visible. Where BPSO, SGPSO, GSA, PSOGSA ans PPSO all can able to classify 102, 103, 105, 107, 109 samples out of 114 samples in cancer test dataset. The accuracy achieved by BPSO, SGPSO, GSA, PSOGSA, PPSO are 89.5, 90.4, 92.1, 93.9, 95.6 respectively. So, it can easily be seen that  PPSO classifier is better than other 4 classifiers.

\par
\begin{figure}[htbp]
\centering
    \subfigure[]{\label{fig:pic1}\includegraphics[width=0.45\linewidth]{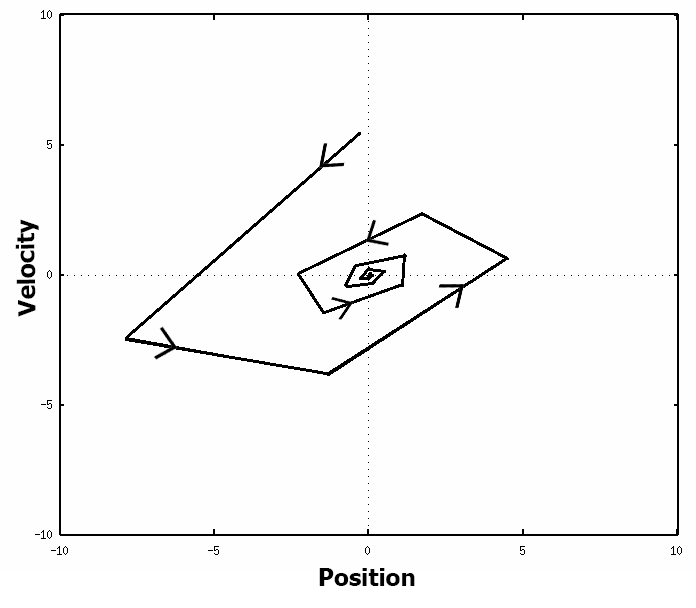}}
    \subfigure[]{\label{fig:pic2}\includegraphics[width=0.45\linewidth]{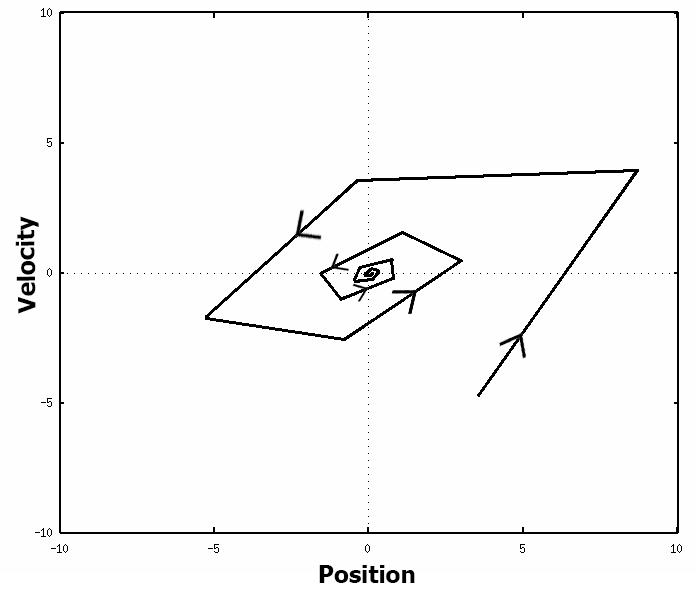}}
  \caption{\ref{fig:pic1} and \ref{fig:pic2} shows the phase trajectory of PPSO with different initial starting point}
  \label{fig:5}
\end{figure}

Figure \ref{fig:5} presents the phase trajectory of the proposed algorithm with different initial points. This experiment has been done using XPPAUT 8.0 \cite{xppaut}. From this phase trajectory, it can be concluded that whatever be the initial position of particle is, proposed inertial weight strategy ($\omega$) will converge to a stable fixed point. It is mentioned that range of $\omega$ must be within the range $0< \omega <\psi -1$.  Hence, it is obvious that, proposed system is stable by analytically as well as experimentally.    

\section{Conclusion}
In this work, a new training algorithm PPSO is introduced and compared with 4 other training algorithms named BPSO, SGPSO, GSA, PSOGSA. To evaluate the performance of the proposed learning algorithm, comparison has been done for classification of real-world benchmark datasets (8 datasets ) taken from UCI machine learning repository. For all benchmark datasets PPSO has shown better performance in terms of convergence rate and avoiding local minima as well as better accuracy. Therefore it can be concluded that the proposed PPSO improves the problem of trapping in local minima with a very good convergence rate. In summary, the results prove that PPSO enhances the problem of trapping in local minima and increases the convergence speed as well as accuracy and can be able to do correct classification more samples as compared to the existing learning algorithms for FFNN. From the stability analysis (theoretically as well as experimentally) the range (i,e $0< \omega <(\psi - 1)$) of inertia weight ($\omega$) is assigned for proposed PPSO is stable. 
\bibliographystyle{named}
\bibliography{mybib}

\end{document}